# Review of Interpretable Machine Learning Models for Disease Prognosis

Jinzhi Shen
Department of Computer and Electrical Engineering
Boston University
Boston, MA, USA
jinzhis9@bu.edu

Ke Ma
Department of Economics
University of California, Santa Cruz
Santa Cruz, CA 95064
kma41@ucsc.edu

*Abstract*— In response to the COVID-19 pandemic, the integration of interpretable machine learning techniques has garnered significant attention, offering transparent and understandable insights crucial for informed clinical decision-making. This literature review delves into the applications of interpretable machine learning in predicting the prognosis of respiratory diseases, particularly focusing on COVID-19 and its implications for future research and clinical practice. We reviewed various machine learning models that are not only capable of incorporating existing clinical domain knowledge but also have the learning capability to explore new information from the data. These models and experiences not only aid in managing the current crisis but also hold promise for addressing future disease outbreaks. By harnessing interpretable machine learning, healthcare systems can enhance their preparedness and response capabilities, thereby improving patient outcomes and mitigating the impact of respiratory diseases in the years to come.

*Keywords— Interpretable machine learning, Domain knowledge, Respiratory diseases, COVID-19, Prognosis*

## I. Introduction

The integration of interpretable machine learning (IML) techniques into healthcare has garnered substantial attention [1]–[8]. The COVID-19 pandemic has underscored the urgent need for transparent and understandable predictive models [1], [3], [7]–[11]. This literature review aims to explore the applications of interpretable machine learning in predicting the prognosis of respiratory diseases, with a specific focus on COVID-19 and its implications for future research and clinical practice.

The COVID-19 pandemic has underscored the urgency of developing accurate prognostic tools to identify high-risk individuals and allocate healthcare resources effectively. In response, numerous interpretable machine learning models have been developed to predict COVID-19 severity [12], [13], mortality risk [14]–[17], and complications [18]. These models not only enhance our understanding of the disease but also offer valuable insights into the broader landscape of respiratory diseases.

Moreover, the experiences gained from applying interpretable machine learning in the context of COVID-19 are poised to have far-reaching implications for future disease outbreaks [19], [20]. By harnessing interpretable machine learning, healthcare systems can improve their preparedness and response capabilities, leading to better patient outcomes and reduced morbidity and mortality from respiratory diseases.

This review will examine the current state of research on interpretable machine learning for respiratory disease prognosis, with a focus on COVID-19. It will explore the various methodologies and applications of interpretable machine learning in this domain, highlighting their strengths, limitations, and potential impact on clinical practice. Additionally, it will discuss future directions and challenges in the field, aiming to inform and guide future research efforts in this critical area of healthcare.

## II. Overview of the Problem Setup

Figure 1. provides a diagram illustrating the problem setup. The status of a patient can be characterized by a variety of electronic healthcare data (EHR) data, including but not limited to 1) the radiological data, such as X-rays and CT scans; (2) symptoms of patients, such as fever, cough, shortness of breath, fatigue, muscle soreness, nausea, and diarrhea; (3) vital signs, such as body temperature, pulse rate, respiration rate, and blood pressure; (4) comorbidities, such as chronic obstructive pulmonary disease, dementia, heart disease, diabetes, cancer, and obesity; (5) blood test measurement, such as white blood cell, hemoglobin, and albumin. Additionally, the temporal information of in-hospital events, such as ICU admission and the use of ventilation, are also important for inferring the disease progression.

The multi-modal measurement contains rich information on COVID-19 patients. After data is collected, both human experts and machine learning models are involved in decision making. Human experts provide valuable domain knowledge and practical experience for evaluating the conditions of patients. The machine learning tools provide a data-driven complementary to derive disease-progression patterns. The physicians-provided domain knowledge and machine-learning identified patterns are further fused to yield an informed decision. Furthermore, the interpretability of machine learning models is enhanced by considering domain knowledge during the model development phase.

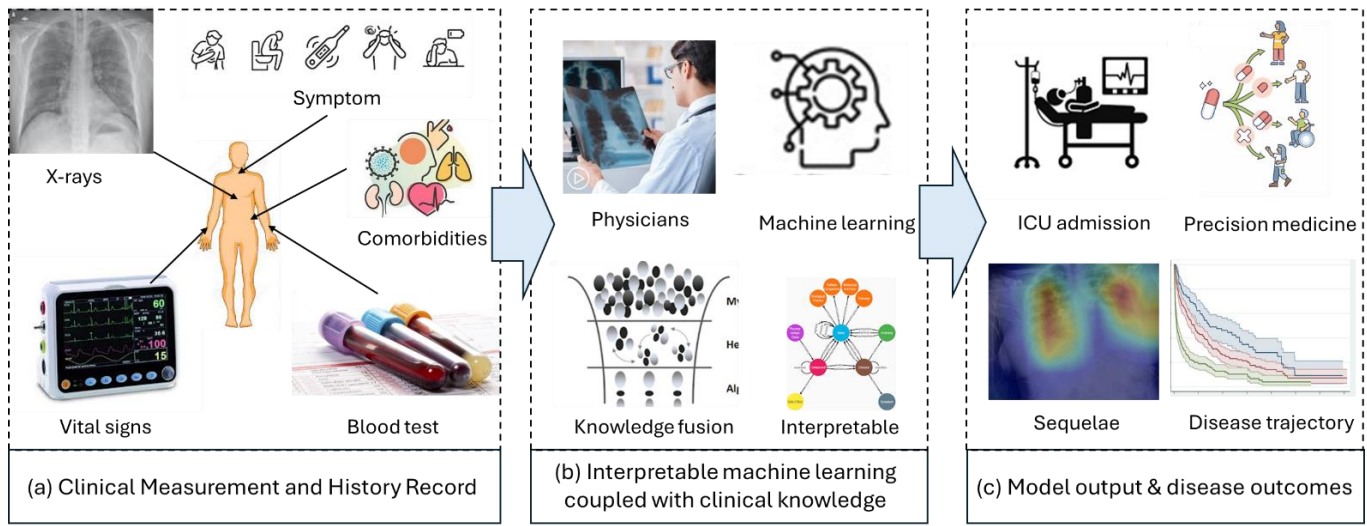

Fig. 1. Problem setup of the machine learning model for COVID-19 prognosis.

The prognosis model can output a variety of predictions. For example, the trajectory of patients can be predicted, with specification on temporal information and uncertainty quantification. The model may predict critical clinical events, such as ICU admission. By anticipating the disease progression in advance, these predictions facilitate precision medicine that prevents severe disease escalation. In a longer temporal frame, some models were able to infer the sequelae after COVID-19 infections.

## III. CLINICAL KNOWLEDGE-INFORMED MODEL DESIGN

The development of interpretable machine-learning models involves a series of steps. Among them, it is important to select relevant features from a complex set of clinical measurements. The underlying reason is that the raw data often contains tens of thousands of features that fluctuate with a variety of physiological conditions [21], [22]. By eliminating the irrelevant information, the noise level is reduced and therefore allows more accurate prognosis.

Besides the input features, the choice of prediction model becomes a key consideration. This choice needs to be made to tradeoff between learning capacity and model generalizability. Commonly known as Occam's razor principle [23], an ideal problem solver shall be represented with a compact set of elements [24]. In practice, the model selection can be made on a validation dataset. For example, by cross-validating the performance of certain models, one can make the comparison and select the best-performing data.

After the model is developed and deployed, it is crucial to update it with additional data or newly derived clinical knowledge. Popular techniques for this include incremental learning, continual learning, and active learning. These techniques enable the model to be actively updated, maintaining its predictive accuracy during the pandemic, especially in response to virus mutations.

### A. Feature selection

Feature selection is crucial in machine learning as it enhances model performance by reducing overfitting, simplifying the model, and improving interpretability. By selecting the most relevant features, models become more robust and efficient, making them faster to train and easier to understand.

In the specific case of COVID-19 prognosis, feature selection helps manage high-dimensional medical datasets and ensures that only the most significant variables are used, thus maintaining data quality and model tractability [25]–[27]. By focusing on the most predictive features, healthcare professionals can gain valuable insights into the disease's progression and severity, leading to more informed clinical practices and effective public health strategies.

There are two ways of selecting features for COVID-19 prognosis: 1) clinical-knowledge-based selection and 2) statistics-based selection.

1) **Clinical-Knowledge-Based Selection**: This approach relies on the expertise of medical professionals and existing clinical knowledge about COVID-19 to identify relevant features. Clinicians and researchers use their understanding of the disease's pathology, symptoms, and risk factors to choose features that are likely to be important for prognosis. For instance, well-known clinical indicators such as age, comorbidities (like hypertension and diabetes), vital signs (such as oxygen saturation and respiratory rate), and specific laboratory values (such as C-reactive protein and D-dimer levels) are often selected based on their established association with COVID-19 outcomes. This method ensures that the selected features have a meaningful and interpretable connection to the disease, facilitating the development of models that are clinically relevant and trusted by healthcare professionals.

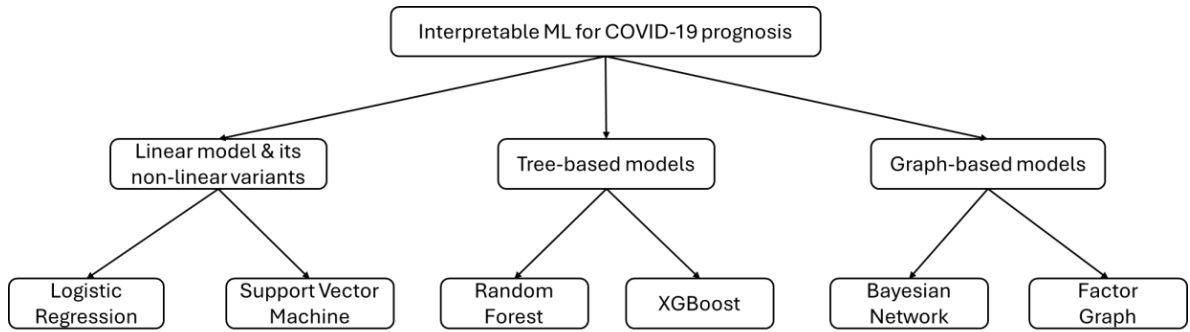

Fig. 2. Taxonomy of interpretable machine learning models for COVID-19 prognosis.

2) **Statistics-Based Selection**: This approach uses statistical methods and machine learning algorithms to automatically identify the most relevant features from a dataset. Techniques such as correlation analysis, mutual information, recursive feature elimination, and regularization methods (like Lasso and Ridge regression) help in selecting features that have the strongest statistical relationship with the target outcome (e.g., disease severity, mortality). Advanced methods like principal component analysis (PCA) and feature importance rankings from tree-based models (such as random forests or gradient boosting machines) are also commonly used. Statistics-based selection can uncover hidden patterns and relationships that might not be immediately apparent to clinicians, and it is particularly useful when dealing with large and complex datasets where manual selection is impractical.

*B. Model consideration*

The choice of model is crucial for COVID-19 prognosis as it directly impacts the accuracy, reliability, and interpretability of predictions. Selecting an appropriate model ensures that it can effectively handle the specific characteristics of COVID-19 data. Fig. 2. summarizes the prevalent interpretable machine learning models for the COVID-19 prognosis into three groups: (1) Linear-summation models and their non-linear variants, (2) tree-based models, and (3) graph-based models.

**Linear-summation models**, such as logistic regression (LR) and support vector machines (SVM), are highly interpretable because they provide clear, quantitative relationships between input features and the predicted outcome [13], [28]–[30]. For instance, LR models developed by Chen et al. predict the need for ventilatory support based on demographics, symptoms, comorbidities, laboratory tests, and vital signs, demonstrating interpretability by allowing clinicians to understand how each factor contributes to the prediction [30], [31]. SVMs, while slightly more complex, still allow for interpretability through feature weights and decision boundaries, as demonstrated in their application for predicting mortality risk based on clinical features like CRP, BUN, serum calcium, albumin, and lactic acid according to Booth et al [28].

**Tree-based models**, such as random forests (RF) and XGBoost, are interpretable due to their inherent structure of decision trees [15], [18], [28]. Ma et al. demonstrated an RF and XGBoost model assessing mortality risk using LDH, CRP, and age, elucidating how different features contribute to predictions through individual decision paths within the trees [15]. Another XGBoost model, highlighted by Yan et al., predicts mortality risk within a 10-day window using LDH, lymphocyte count, and hs-CRP, providing valuable insights into patient prognosis through visualized feature importances and decision rules[1].

**Graph-based models**, such as Bayesian networks [32] and factor graphs [33], capture and represent dependencies and conditional probabilities between features in a graphical form. Bernaola et al. utilized a Bayesian network model to visualize interconnected factors like hospital stay duration, ICU admission, and mortality risk, providing actionable insights for patient care [32]. Similarly, Cao et al. employed a factor graph to predict ICU admission based on comorbidities, laboratory tests, and vitals, offering a structured view of how these variables collectively influence patient outcomes[33]. Additionally, Wang et al. developed a graph-based model to analyze CT image features, achieving high accuracy in predicting disease aggravation among patients [34].

*C. Update strategy*

The dynamic and evolving nature of the COVID-19 pandemic necessitates the continuous updating of machine learning models to maintain their accuracy, reliability, and relevance. As new data becomes available and our understanding of the disease deepens, static models quickly become outdated, potentially leading to inaccurate predictions and suboptimal clinical decisions. The complexity and variability of COVID-19, as highlighted by the rapid emergence of new variants and shifting patterns in patient outcomes, further underscores the need for models that can adapt and evolve in real-time. There are mainly three types of strategies for updating the machine learning models using newly available data.

**Incremental Learning:** Incremental learning approaches, as proposed in the second paper, allow models to be updated continuously as new data becomes available. This method, as developed in [35], [36], involves dynamically adding new models or updating existing ones to reflect the latest data trends. By using a dynamic ensemble method based on a bagging scheme, models can maintain high predictive performance while adapting to new information. This approach ensures that the model remains relevant and accurate over time.

TABLE I
A LIST OF INTERPRETABLE MACHINE LEARNING METHODS DEVELOPED FOR COVID-19 PROGNOSIS

| Machine Learning Models | Input clinical features | Model Output | Patient number | Performannces | Refs |
|---|---|---|---|---|---|
| RF and XGboost | LDH, CRP, age | Mortality risk | 205 | AUROC: 0.9667 | [15] |
| SVM | CRP, BUN, serum calcium, ALB, and lactic acid | Mortality risk | 398 | AUROC: 0.93 AUPRC: 0.76 | [28] |
| XGBoost | LDH, lymphocyte and hs-CRP | Mortality risk in a 10-days' time window | 485 | Accuracy: 81% | [1] |
| Bayesian networks | Time of stay in the hospital, ICU admission and death | Mortality risk | 1,645 | Accuracy: 86.4% | [32] |
| Factor graph | Comorbidities, laboratory tests, and vitals | ICU admission | 533 | AUC: 0.87 | [33] |
| LR, SVM, DT, RF, XGBoost | 7 first-order features, 3 shape features, 9 texture features of CT images | Aggravation Possibility | 188 | Best AUC: 0.843 | [34] |
| Stacking ensemble | Phone-surveyed data, including but not limited to demographics, recovery days, and symptoms during disease | Risk of heart disease | 180 | Accuracy: 93.11% Specificity: 94.81% Precision: 95.14% Recall：91.38% | [40] |
| LR | Demographics, symptoms, comorbidities, laboratory tests, and vital signs | Ventilatory support | 301 | AUC: 0.823 | [30] |
| XGBoost | clinical data, such as sex, age, main clinical symptoms and radiological images | Prediction of pulmonary fibrosis and normal lungs | 1175 | Accuracy 99.37±0.83 Precision : 99.54±0.35 Recall: 99.85±0.22 Specificity: 99.94±0.26 | [18] |
| LR, RF, and SVM | demographic variables & symptoms | predictive value of smell and taste disorders | 777 | LR: AUC 0.759 RF: AUC 0.777 SVM: AUC 0.722 | [13] |

CRP: C-reactive protein
hs-CRP: high-sensitivity C-reactive protein
LDH: lactate dehydrogenase
BUN: blood urea nitrogen
ALB: albumin

**Human-in-the-Loop Systems:** Incorporating human expertise in the model updating process can enhance the relevance and accuracy of predictions. A reinforcement learning-based human-in-the-loop system, as proposed in [37], can continuously learn from clinician feedback and new clinical data, refining its predictions to better support clinical decision-making. This approach leverages the strengths of both machine learning and expert human judgment.

**Active Learning:** Active learning strategies involve selectively updating models based on the most informative new data points, thereby improving efficiency and accuracy. Additionally, using cross-population train/test models ensures that the models generalize well across different demographic and geographic populations. This approach, as discussed in the [38][39], is crucial for handling the diverse and multimodal data associated with COVID-19, enabling more robust and generalizable predictions.

## IV. IMPLCATION FOR HEALTHCARE SYSTEM

Interpretable machine learning not only offers trustworthiness to practitioners but also demonstrates effectiveness in prognosis performance. In this section, we will first review the performance of these interpretable machine learning models during the COVID-19 pandemic. Subsequently, we will explore opportunities to extend this work to future efforts in combating diseases.

### A. Quantified effectiveness for COVID-19 pandemic

These interpretable machine learning models achieve high prediction performance in multi-center clinical cohorts, demonstrating their robustness and applicability across diverse patient populations. One notable approach utilized RF and XGBoost models to assess mortality risk based on clinical features such as lactate dehydrogenase (LDH), C-reactive protein (CRP), and age. This method, evaluated in a cohort of 205 patients, achieved an exceptional area under the receiver operating characteristic curve (AUROC) of 0.9667, demonstrating its high discriminatory power and potential for guiding clinical interventions effectively [15].

SVM models have also shown significant efficacy. Booth et al. implemented an SVM model to predict mortality risk using features including CRP, blood urea nitrogen (BUN), serum calcium, albumin (ALB), and lactic acid levels. This study, involving 398 patients, reported an AUROC of 0.93 and an area under the precision-recall curve (AUPRC) of 0.76 [28]. These metrics reflect the robust capability of SVM in identifying high-risk patients, crucial for timely and effective intervention. Similarly, Yan et al. used XGBoost to predict mortality risk within a 10-day window, leveraging features such as LDH, lymphocyte count, and high-sensitivity CRP (hs-CRP). The model, tested on 485 patients, achieved an accuracy of 81%, highlighting its utility in short-term prognostication and resource allocation during critical periods [1].

Further advancing the field, Bayesian networks have been applied to predict mortality risk based on hospital stay duration, ICU admission, and death records. Bernaola et al. achieved an accuracy of 86.4% in a cohort of 1,645 patients, underscoring the model's effectiveness in managing hospital resources and developing targeted care strategies [32]. Complementing this, Cao et al. used factor graphs to predict ICU admission based on comorbidities, laboratory tests, and vital signs in a study involving 533 patients. The model's AUC of 0.87 demonstrates its potential in predicting critical care needs, essential for ensuring appropriate care levels and optimizing the use of intensive care resources [33].

Machine learning models have also extended their utility to imaging and long-term health risk prediction. Wang et al. combined logistic regression, SVM, decision tree, random forest, and XGBoost models to analyze CT image features [34], achieving an AUC of 0.843 in predicting disease aggravation among 188 patients. Additionally, Gupta et al. used a stacking ensemble model based on phone-surveyed data to predict post-COVID-19 heart disease risk, achieving impressive metrics such as 93.11% accuracy and 94.81% specificity [40]. These studies illustrate the versatility and effectiveness of ML models in addressing both immediate and long-term health outcomes, thereby enhancing patient management and optimizing healthcare resource allocation during the pandemic and beyond.

### B. Furture implication

The success of interpretable deep learning models during the COVID-19 pandemic has opened numerous opportunities to extend their application to future efforts in combating diseases. One significant opportunity lies in the realm of early disease detection and diagnosis. By leveraging the ability of these models to process vast amounts of data from various sources, including medical imaging, genetic information, and electronic health records, it is possible to identify patterns and markers indicative of early disease stages. This can lead to timely interventions and improved patient outcomes.

Another promising area is personalized medicine. Interpretable deep learning models can analyze individual patient data to predict responses to different treatments, allowing for the customization of therapeutic approaches. This personalized approach not only enhances treatment efficacy but also minimizes adverse effects, thereby improving the overall quality of healthcare. For instance, in oncology, these models can help tailor chemotherapy regimens based on a patient's genetic profile and tumor characteristics.

Moreover, the ability to update and refine these models with new data ensures that they remain relevant and accurate over time. This continuous learning capability is particularly crucial in dealing with emerging infectious diseases and evolving pathogens. By incorporating real-time data from ongoing research and clinical practice, interpretable deep learning models can adapt to new strains of viruses or bacteria, providing up-to-date insights and guidance for public health responses.

In the context of global health, these models can be deployed in resource-limited settings to improve disease surveillance and control. By utilizing interpretable models, healthcare workers in these regions can make informed decisions even with limited access to advanced medical infrastructure. This democratization of advanced healthcare technologies can play a vital role in addressing health disparities and improving outcomes in underserved populations.

Furthermore, the transparency and explainability of these models foster trust among healthcare professionals and patients, facilitating their integration into clinical workflows. This trust is essential for the widespread adoption of AI-driven tools in healthcare, as it allows practitioners to understand and verify the reasoning behind model predictions and recommendations. This, in turn, can lead to better-informed clinical decisions and enhanced patient care.

## V. Conclusion

In summary, this paper reviews the pivotal role of interpretable machine learning in facilitating collaborative decision-making between healthcare professionals and machine learning algorithms. These models offer transparent insights into disease progression, treatment responses, and mortality risks, empowering healthcare teams to optimize patient care and resource allocation. Amidst the urgency of the COVID-19 pandemic, interpretable machine learning has proven indispensable in predicting disease severity and guiding clinical management. Moving forward, sustained research efforts are essential to address challenges and ensure the seamless integration of interpretable machine learning into routine clinical practice. Ultimately, this approach holds promise for improving patient outcomes and mitigating the impact of diseases such as COVID-19 on a global scale.